\documentclass[conference,letterpaper]{IEEEtran}
\usepackage[letterpaper, left=0.75in, right=0.75in, bottom=0.75in, top=0.75in]{geometry}
\def\IEEEtitletopspace{18pt}
\IEEEoverridecommandlockouts
\usepackage{array}
\usepackage{hyperref}
\usepackage{cite}
\usepackage{amsmath,amssymb,amsfonts}
\usepackage{algorithmic}
\usepackage{comment}
\usepackage{graphicx}
\usepackage{adjustbox}
\usepackage{textcomp}
\usepackage{breqn}
\usepackage{amsmath}
\usepackage{xcolor}
\usepackage{booktabs}
\usepackage{xcolor}
\usepackage{subcaption}
\usepackage{caption}
\newcount\Comments  
\usepackage{color}
\definecolor{darkgreen}{rgb}{0,0.5,0}
\definecolor{purple}{rgb}{1,0,1}
\definecolor{teal}{rgb}{0,0.4627,0.5804}
\newcommand{\kibitz}[2]{\ifnum\Comments=1\textcolor{#1}{#2}\fi}

\def\BibTeX{{\rm B\kern-.05em{\sc i\kern-.025em b}\kern-.08em
    T\kern-.1667em\lower.7ex\hbox{E}\kern-.125emX}}
\begin{document}

\title{Deep Reinforcement Learning for Real-Time Ground
Delay Program Revision and Corresponding Flight
Delay Assignments\\

%


\thanks{
$^{1}$K. Liu, F. Hu, and G. Su are with the University of California at Berkeley, Berkeley, CA 94702, USA.

$^{2}$H. Lin is with Northwestern University, Evanston, IL, 60208, USA

$^{3}$X. Cheng is with the University of Illinois at Chicago, Chicago, IL 60607, USA

$^{4}$J. Chen is with the University of British Columbia, Vancouver, BC, Canada V6T 1Z2.

$^{5}$J. Song is with the University of Toronto, Toronto, Ontario M5S 1A1 Canada.

$^{6}$S. Feng is with the Hong Kong University of Science and Technology, Clear Water Bay, Hong Kong.

$^{7}$C. Zhu (corresponding author) is with the Tsinghua University, Beijing, 100084, P. R. China.
}
}

\author{Ke Liu$^{1}$, Fan Hu$^{1}$, Hui Lin$^{2}$, Xi Cheng$^{3}$, Jianan Chen$^{4}$, Jilin Song$^{5}$, Siyuan Feng$^{6}$, Gaofeng Su$^{1}$ Chen Zhu$^{7,*}$}

\maketitle

\begin{abstract}
This paper explores the optimization of Ground Delay Programs (GDP), a prevalent Traffic Management Initiative used in Air Traffic Management (ATM) to reconcile capacity and demand discrepancies at airports. Employing Reinforcement Learning (RL) to manage the inherent uncertainties in the national airspace system—such as weather variability, fluctuating flight demands, and airport arrival rates—we developed two RL models: Behavioral Cloning (BC) and Conservative Q-Learning (CQL). These models are designed to enhance GDP efficiency by utilizing a sophisticated reward function that integrates ground and airborne delays and terminal area congestion. We constructed a simulated single-airport environment, $SAGDP\_ENV$, which incorporates real operational data along with predicted uncertainties to facilitate realistic decision-making scenarios. Utilizing the whole year 2019 data from Newark Liberty International Airport (EWR), our models aimed to preemptively set airport program rates. Despite thorough modeling and simulation, initial outcomes indicated that the models struggled to learn effectively, attributed potentially to oversimplified environmental assumptions. This paper discusses the challenges encountered, evaluates the models' performance against actual operational data, and outlines future directions to refine RL applications in ATM.

\end{abstract}


\section{Introduction}
Over the past four decades, U.S. air traffic demand has increased more than sixfold, with an anticipated average growth rate of 2.2\% over the next twenty years \cite{forecast2016fiscal}; \cite{campbell2017}; \cite{liu2023airborne}. This escalating demand places additional strain on National Airspace System (NAS) components such as airports and sectors, and increases the workload for traffic controllers and managers. U.S. Air Traffic Management (ATM) encompasses Air Traffic Control (ATC) and the Traffic Flow Management System (TFMS) \cite{liu2021miles}. ATC coordinates the separation of aircraft through controller and pilot collaboration, while traffic flow managers employ various Traffic Management Initiatives (TMIs) to align traffic demand with capacity, strategically relocating delays to locations where they can be absorbed safely and efficiently. These TMIs, which have evolved significantly over decades, include diverse concepts, procedures, automation, and control characteristics.

One of the most common TMIs used today is Ground Delay Program (GDP). A GDP is necessary and will be activated when anticipated traffic excessively exceeds capacity at an arrival airport—often due to forecasted adverse weather—GDP assigns ground delays at departure airports, effectively transferring airborne delays to the ground. A typical GDP is characterized by four parameters:

\begin{itemize}
    \item \noindent Program rate (PAAR): the maximum quarterly arrival rate for flights at restricted airport.

    \item \noindent Scope: the set of origin airports whose departures are subject to delays by the GDP. The scope is usually defined as great circle distance or control centers. 

    \item \noindent Release time: the time when GDP is modeled and released to the in-scope origin airports and corresponding restricted flights 

    \item \noindent Duration: the time window of GDP operation expressed in local time at the destination airport which also includes the start and end time of the program. 
\end{itemize}
    
In practice, GDP design is largely manual, aided by the Flight Schedule Monitor, a decision-support tool \cite{cox2016ground};\cite{mukherjee2014predicting};\cite{liu2024excess}. Traffic managers set the scope, duration, and program rate—reflecting airport capacity—which are then entered into the monitor to calculate ground delays for affected flights according to scheduling principles. Additionaly, GDP, a strategic Traffic Management Initiative (TMI), requires planning several hours in advance based on anticipated capacity and demand. However, these forecasts often fail to capture system uncertainties, leading to potential GDP revisions during operation if conditions change. This results in multiple GDP events, such as new issuance, revisions, or cancellations. 

Improvement opportunities arise primarily from two types of uncertainties: supply (e.g., GDP program rate and time window) and demand (e.g., flight time variations and airline operational changes like cancellations and substitutions). There is a fundamental trade-off between control accuracy and the reliability of information over time. Initiating GDP earlier allows more flights to receive ground delays but increases the uncertainty in capacity and flight times. Despite these complexities, current real-time GDP designs remain suboptimal and require further enhancement \cite{bloem2015ground}, \cite{cheng2024using}. Limited number of research focuses on optimizing GDP design and execution. Objective functions usually involve the weighted sum of ground and air delay, and airline equity with constraints from uncertainties regarding flight delays, weather conditions and airport capacity \cite{bloem2015ground}; \cite{yilmaz2021deep}; \cite{george2015reinforcement},\cite{cox2016ground}. Various solvers such as common machine learning, dynamic programming \cite{cheng2024carsharing,cheng2024electric}, stochastic optimization \cite{cheng2024autonomous,cheng2023estimating}, and reinforcement learning, were applied to find optimal GDP parameters and to predict flight performance \cite{ganesan2010predicting} \cite{kalliguddi2017predictive} \cite{deshpande2012impact}. Due to the great complexity in the NAS and technological limitations, such models have yet to reach the stage of practical application.

Against this background, this paper aims to optimize GDP operations in the NAS by developing and applying a time-sequential-based method that not only shifts delays without reducing throughput or exacerbating delays but also enhances overall system performance. Specifically, we will design a deep reinforcement learning (RL) model that utilizes real-time weather data and both realized and predicted flight delays. This model will dynamically adjust GDP parameters and allocate ground delays to affected flights, maximizing efficiency across multiple dimensions including total flight delay, airport utilization, airline equity, and controller workload.

\nocite{bloem2015ground}
\nocite{ganesan2010predicting}
\nocite{mukherjee2014predicting}
\nocite{george2015reinforcement}
\nocite{cox2016ground}
\nocite{zhu2017communication}
\nocite{kalliguddi2017predictive}
\nocite{yilmaz2021deep}
\nocite{deshpande2012impact}

\section{Experiment Setup} 
\subsection{Data Sources and Feature Extraction}

Our study utilized three comprehensive datasets collected throughout 2019 from January $1^{st}$ to December $31^{st}$ of 2019 for the national airspace system (NAS) in the U.S. 

\begin{enumerate}
  \item \emph{ASPM Flight Level Data.} This dataset encompasses records from commercial, general aviation, air taxi, and military flights at 77 major U.S. airports. It includes detailed information on scheduled and actual departure and arrival times, as well as delays for each flight.
  
  \item \emph{ASPM Airport Quarter Hour Data.} This dataset provides operational and meteorological data recorded every 15 minutes at the top 77 commercial airports. It includes metrics such as acceptance rates and weather conditions—temperature, ceiling, wind speed, and visibility.
  
  \item \emph{ATCSCC GDP Advisory Database.} Captures all Ground Delay Programs (GDPs) issued within the National Airspace System (NAS) in 2019. Essential details include the targeted airport, time of issuance, duration, reasons for the GDP, program rates, affected scopes, and any exemptions. 

\end{enumerate}

Our analysis identified 33 airports with active GDP operations, with Newark (EWR), San Francisco (SFO), and LaGuardia (LGA) issuing the most programs. Each of these airports released at least 15 GDPs throughout the year. The accompanying upper bar plot visualizes the number of GDPs issued from these airports, while the lower graph displays the total number of flights affected by these GDPs per airport.

We established the following criteria to determine whether a flight was impacted by a GDP:

\begin{itemize}
    \item[a.] if its flight schedule arrival time is within the GDP effective time window; 
    
    \item[b.] if its departure airport is in the restricted control centers/ within the distance scope;

    \item[c.] the flight has not taken off when the program is modeled and released.
\end{itemize}

\begin{figure*}[ht]
\centering
\includegraphics[width=1.9\columnwidth]{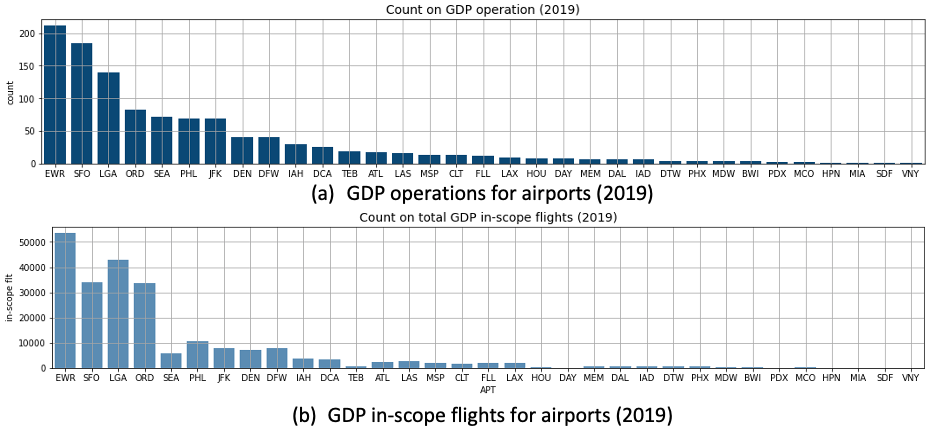} 
\caption{Statistics for GDP operation in 2019}
\label{fig:stats}
\end{figure*}

If a, b and c are all satisfied, the flight is restricted by the given GDP event. We applied this detection method to every GDP operation in NAS during 2019, and Figure \ref{fig:stats} shows the total number of restricted flights for each airport. This feature is included in the next Scenario Generation for the RL model. 

\subsection{Scenario Generation} 
In this project, we develop a model targeting the management of single-airport ground delay programs (GDP). Newark Liberty International Airport (EWR), which issued the most GDPs in 2019 (215 in total), serves as our primary case study. Our model simulates operations during daytime hours from 10:00 UTC to 06:00 UTC the following day, equating to 6:00 AM to 2:00 AM EDT or 7:00 AM to 3:00 AM EST. Each 15-minute interval within this window represents a state, totaling 80 states per trajectory.

\begin{enumerate}
    \item \noindent \emph{Action}:\\ to set the program arrival rate (PAAR) for the given arrival airport for the next $n$ intervals in time $t$. The $i^{th}$ entry of action is the PAR chosen in $t$ for time $t+i$. Currently, we select $n = 8$, so an action corresponds to the PAR for the next $2$ hours.

    \item \noindent \emph{State}: \\ the state at time $t$ ($S_t$, during $15$ min) includes current operational conditions, weather information and flight schedule: 
    \vspace{-0.4em}
    \begin{itemize}
        \item ARR\_RATE: airport-specific actual capacity at time $t$ (variable dimension = 1)
        \item ACT\_ARR: airport-specific actual arrival flights at time $t$ (variable dimension = 1)
        \item CEILING: airport-specific time-of-day discretized ($k = 1, 2, 3, 4$) ceiling; intervals are [0, 5], (5, 10], (10, 30], (30, 100] (100 feet) for the next 8 quarter hours (variable dimension = $1 \times 8$)
        \item WIND\_A: airport-specific specific time-of-day wind angle (range from 0 to 359) for the next 8 quarter hours (variable dimension = $1 \times 8$)
        \item WIND\_V: airport-specific specific time-of-day wind speed for the next 8 quarter hours (variable dimension = $1 \times 8$) 
        \item VISIBLE: airport-specific time-of-day discretized ($k = 1, 2, 3, 4$) visibility; intervals are [0, 1], (1, 3], (3, 5], (5, 10] (statute miles) for the next 8 quarter hours (variable dimension = $1 \times 8$)
        \item RW\_NUM: airport-specific number of runways in operation for the next 8 quarter hours (variable dimension = $1 \times 8$)
        \item ARR\_DEMAND: the number of flights which plan to arrive at destination airport for the next 8 quarter hours (variable dimension = $1 \times 8$)
        \item DEP\_DEMAND: the number of flights which plan to take off from the destination airport for the next 8 quarter hours (variable dimension = $1 \times 8$)
        \item ENROUTE\_FLT: the number of enroute flight matrix whose $(i, j)$ entry represents the number of flights enroute at time $i$ arriving at time $j$. It is a upper triangular matrix (variable dimension = $8 \times 8$).
    \end{itemize}
    In all, there are 122 variables to describe the state for one observation.\\

    \item \noindent \emph{Delay and Reward Function}: \\ reward at time t is defined as the negative sum of the ground delay and airborne delay of the arrival flights for the next n intervals in time t, and penalty for airspace congestion level with benchmark of 10 holding flights. As the airborne delay is more costly than ground delay and extra airborne delays can be unsafe, we assign different unit cost for air delay and arrival delay.  The reward at time t is defined as

    $$r_t = - \{qtr \times \sum_{i=1}^{n} [ c_{gnd} \times GD_{t+i}        $$
    $$\hspace*{1.5cm}+ c_{air} \times AD_{t+i} ] + p \times (NH_{t}-10)\}$$

    where \:$GD_{t+i} = $ the number of ground-delayed flights at time $t+i$,\\
    \hspace*{1.1cm}  $AD_{t+i} = $ the number of airborne delayed of flights at time $t+i$, \\
    \hspace*{1.1cm}  $NH_{t} = $ the number of air-holding flights near terminal area at time $t$,\\
    \hspace*{1.2cm}  $qtr = $ the time interval for each model step which 15 min in the model,\\
    \hspace*{1.1cm}  $c_{gnd} = $ the cost of one unit of ground delay which is 1 per min,\\
    \hspace*{1.1cm}  $c_{air} = $ the cost of one unit of ground delay which is 2.5 per min,\\
    \hspace*{1.1cm}  $p  = $ the penalty of congestion due to holding flights which is 10 per flight.\\

    \item \noindent \emph{State Transition}: \\ Given the action $a_t$ and $obs_t$, we developed a environment ($SAGDP\_ENV$) to generate next state at time $t+1$. At initial stage, $SAGDP\_ENV$ input information from given the datasets including actual airport capacity ($ARR\_RATE$), weather and operational conditions ($CEILING$, $WIND\_A$, $WIND\_V$, $VISIBLE$, $RW\_NUM$), and flight arrival and departure schedule for the entire trajectory (20 hrs = 80 quarter hours), and set the default $ACT\_ARR$, $ARR\_DEMAND$ list, $DEP\_DEMAND$ list and $ENROUTE\_FLT$ matrix. At each step, $SAGDP\_ENV$ will
    
    \begin{itemize}
    \item[(1)] take the airport landing capacity, $ARR\_RATE$, at time $t$;
    \item[(2)] take the airport conditions, $CEILING$, $WIND\_A$, $WIND\_V$, $VISIBLE$, $RW\_NUM$, and $DEP\_DEMAND$ from time $t$ to $(t+8)$;
    \item[(3)] apply the queuing diagram based on ration-by-schedule and first-come-first-served rults, with inputs of $a_t$, $ARR\_DEMAND_{t-1}$, and $ENROUTE\_FLT_{t-1}$, to assign the ground delay to the detected in-scope flights in the following 8 quarter hours as $GD_{t}$ to $GD_{t+8}$;
    \item[(4)] update $ACT\_ARR_t$, $ARR\_DEMAND_{t}$, and $ENROUTE\_FLT_{t}$;
    \item[(5)]calculate $AD_{t}$ and $NH_{t}$.
    \end{itemize}
\end{enumerate}


\subsection{TFM Agent}
\begin{figure*} [ht]
\centering
\hspace*{-0in}
\begin{subfigure}{.5\textwidth}
  \centering
  \includegraphics[width=1.0\linewidth]{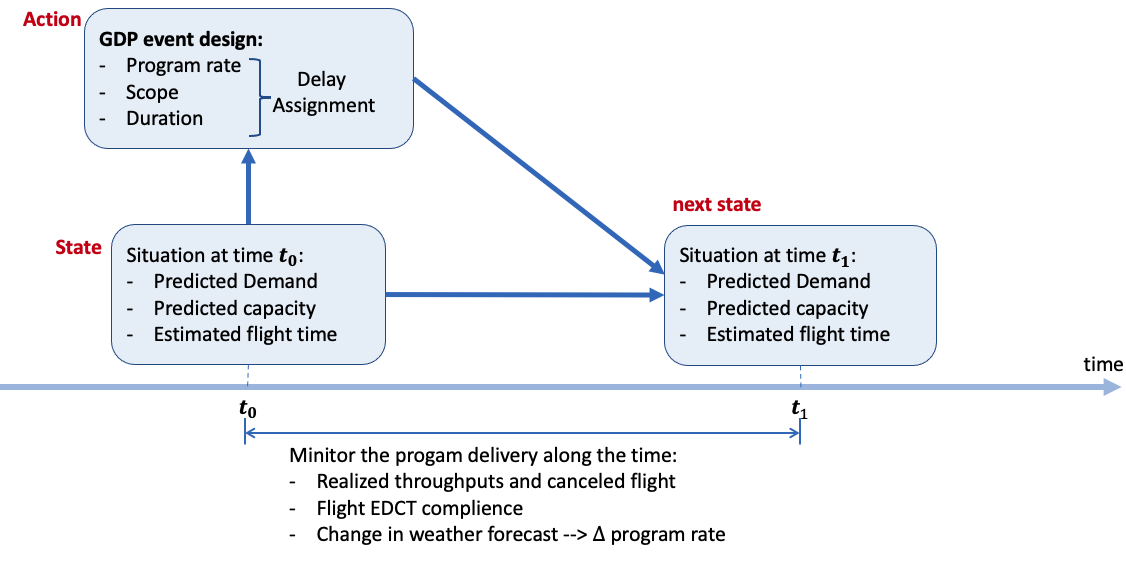}
  \caption{Initiation}
  \label{fig:framework_1}
\end{subfigure}%
\hspace*{0in}
\begin{subfigure}{.5\textwidth}
  \centering
  \includegraphics[width=1.0\linewidth]{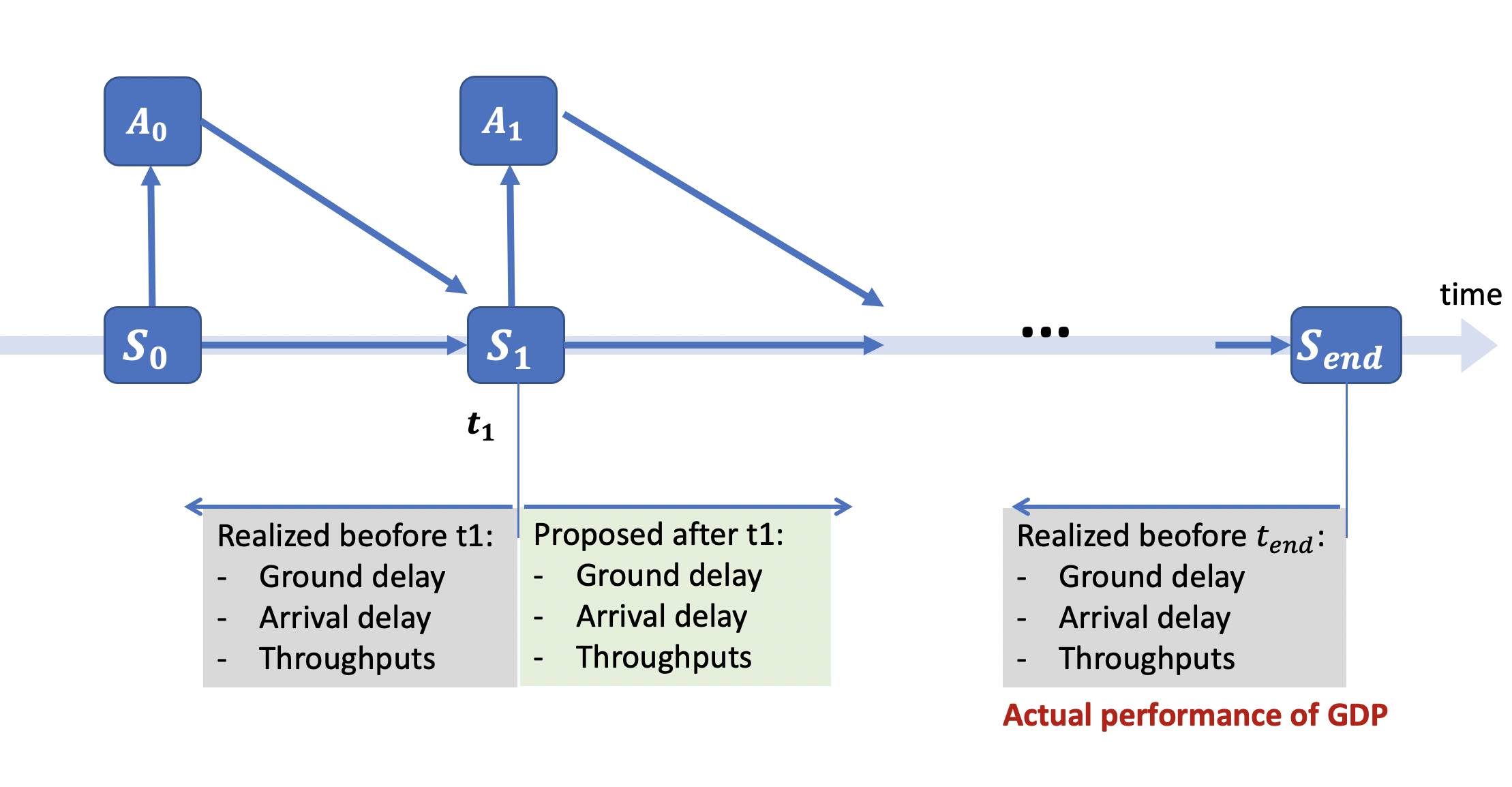}
  \caption{Revisions}
  \label{fig:framework_2}
\end{subfigure}
\caption{Decision Process for GDP Initiation and Revisions.}
\label{fig:framework}
\end{figure*}

\noindent In the field of reinforcement learning, the agent interacts with the environment and accordingly evaluates and selects the best action (i.e., GDP) for implementation in order to maximize a reward, which in our case corresponds to reducing the negative impacts of airborne delays. Even though the best action at time $t$ takes into account the \emph{predicted} future conditions (e.g., the downstream GDP could be influenced by the future observed demand), the time $t$ model is only used to select the time $t$ action. As shown in Figure \ref{fig:framework}, the agent's action at time $t_i$ results in an updated airport state at time $t_{i+1}$.

We choose Behavioral Cloning (BC) and Conservative Q-Learning (CQL, \cite{kumar2020conservative}) as our Traffic Flow Management (TFM) agent. BC is one of the simplest forms of imitation learning, which aims at learning the expert’s policy via supervised learning. It is simple and easy to implement, which makes it a good baseline of our offline RL approach. CQL is a SAC-based data-driven DRL algorithm, which achieves state-of-the-art performance in offline RL problems. CQL mitigates overestimation error by minimizing action-values under the current policy and maximizing values under data distribution for underestimation issue. In CQL, the standard temporal-difference (TD) error objective is augmented a simple Q-value regularizer weighted by a hyperparameter $\alpha$ ($\alpha = 0 \Rightarrow $ DQN), which makes it straightforward to implement on top of existing DRL implementations. 


\section{Experiment Results}


\begin{figure*} [ht]
\centering
\hspace*{-0in}
\begin{subfigure}{.5\textwidth}
  \centering
  \includegraphics[width=1.0\linewidth]{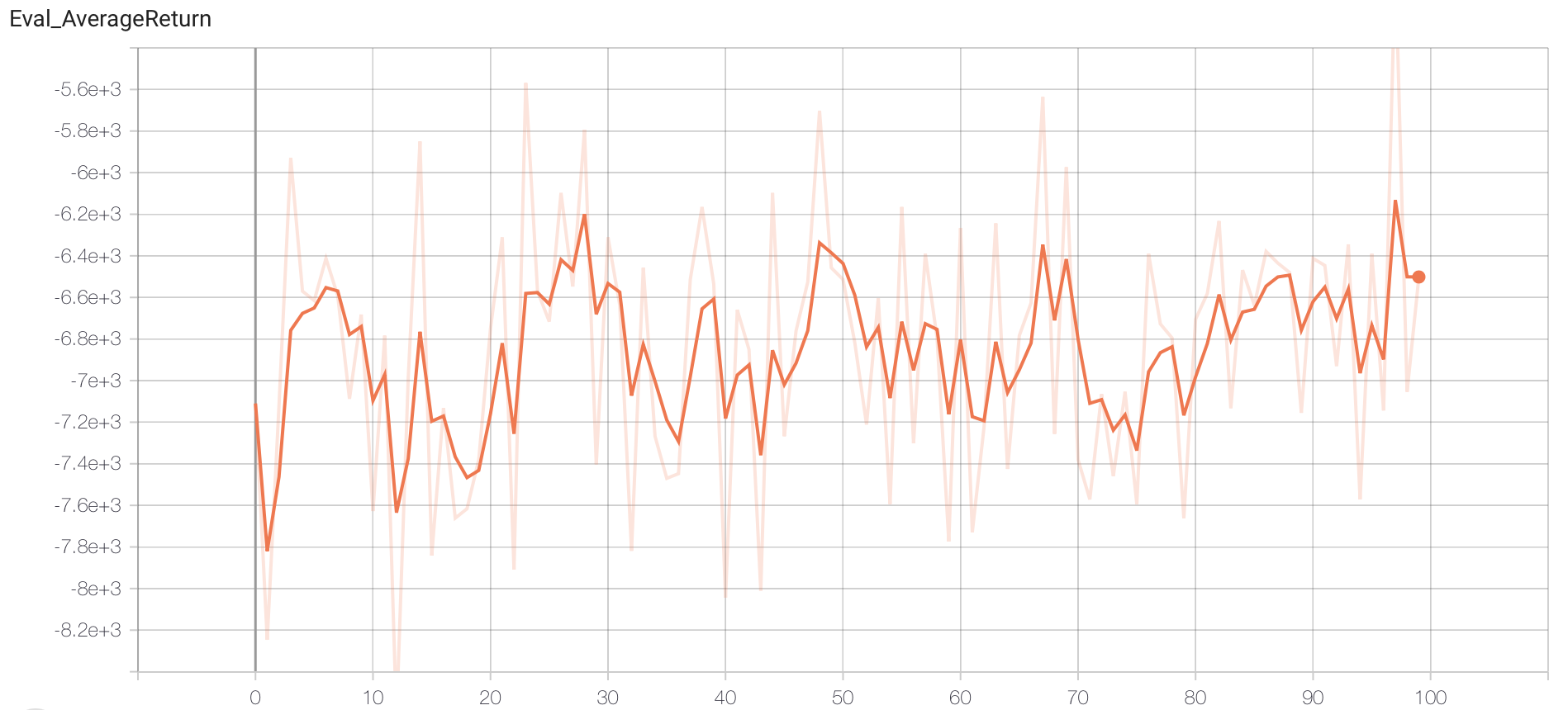}
  \caption{\texttt{eval\_batch\_size = 100}}
  \label{fig:exp1}
\end{subfigure}%
\hspace*{0in}
\begin{subfigure}{.5\textwidth}
  \centering
  \includegraphics[width=1.0\linewidth]{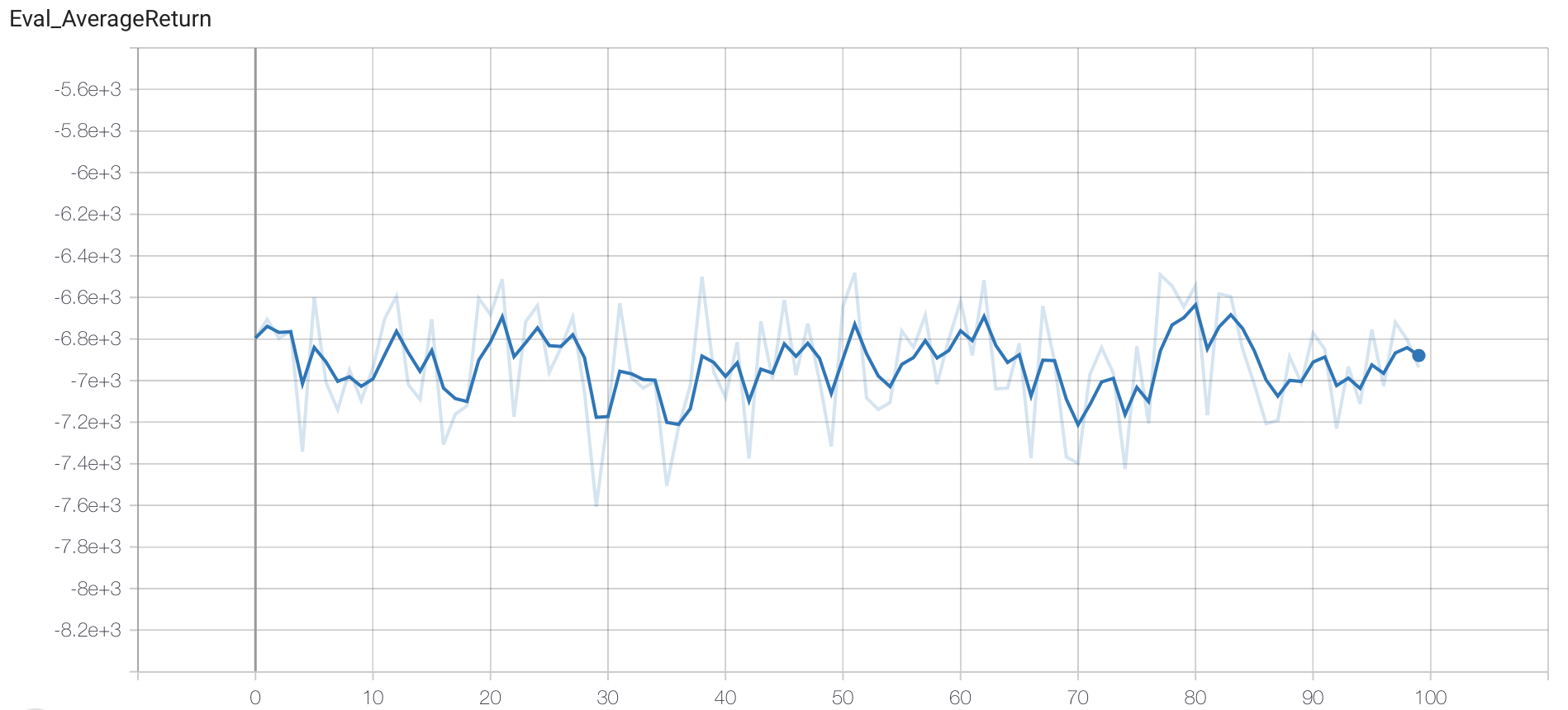}
  \caption{\texttt{eval\_batch\_size = 1000}}
  \label{fig:exp2}
\end{subfigure}
\caption{Performance of BC agent.}
\label{fig:exp_BC}
\end{figure*}

\begin{figure*} [ht]
\centering
\hspace*{-0in}
\begin{subfigure}{.5\textwidth}
  \centering
  \includegraphics[width=1.0\linewidth]{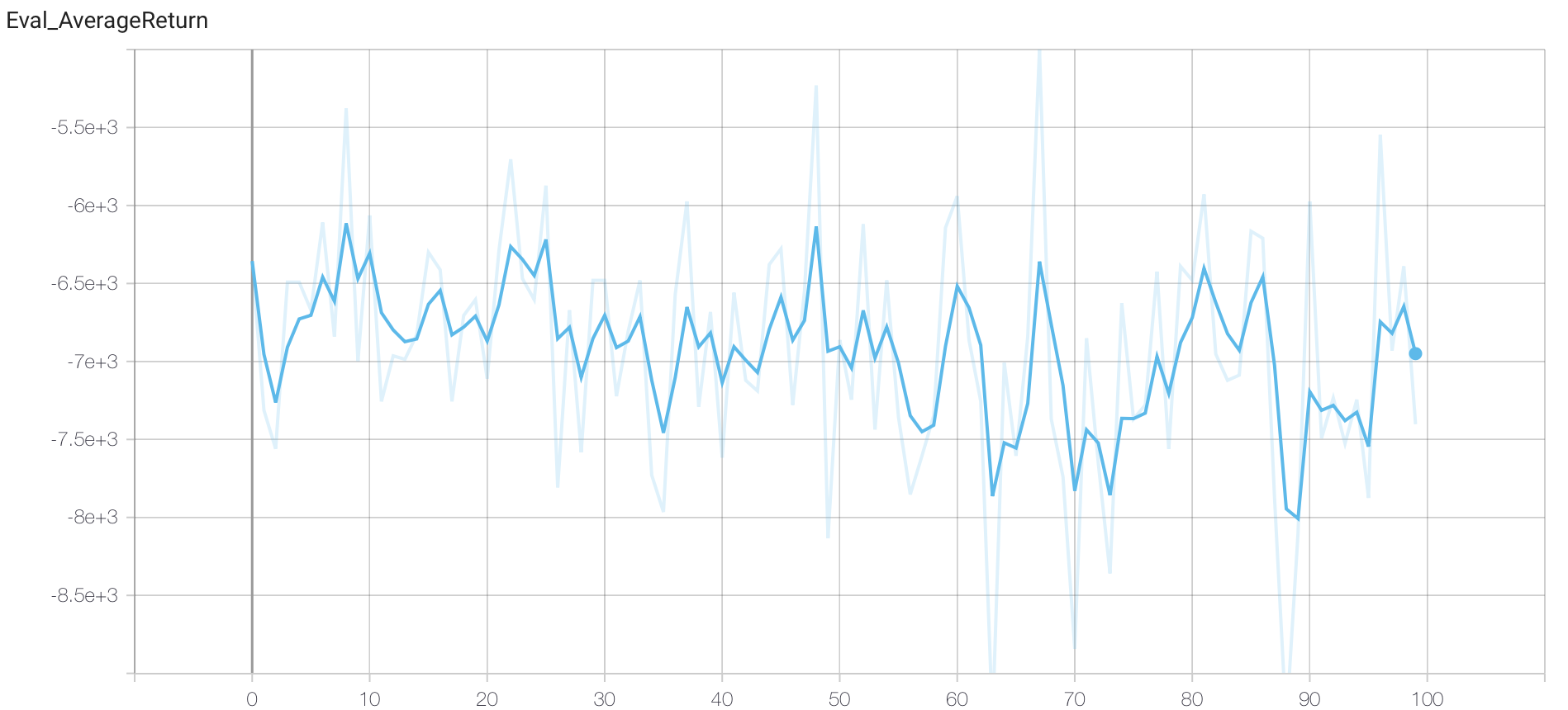}
  \caption{\texttt{eval\_batch\_size = 100}}
  \label{fig:exp3}
\end{subfigure}%
\hspace*{0in}
\begin{subfigure}{.5\textwidth}
  \centering
  \includegraphics[width=1.0\linewidth]{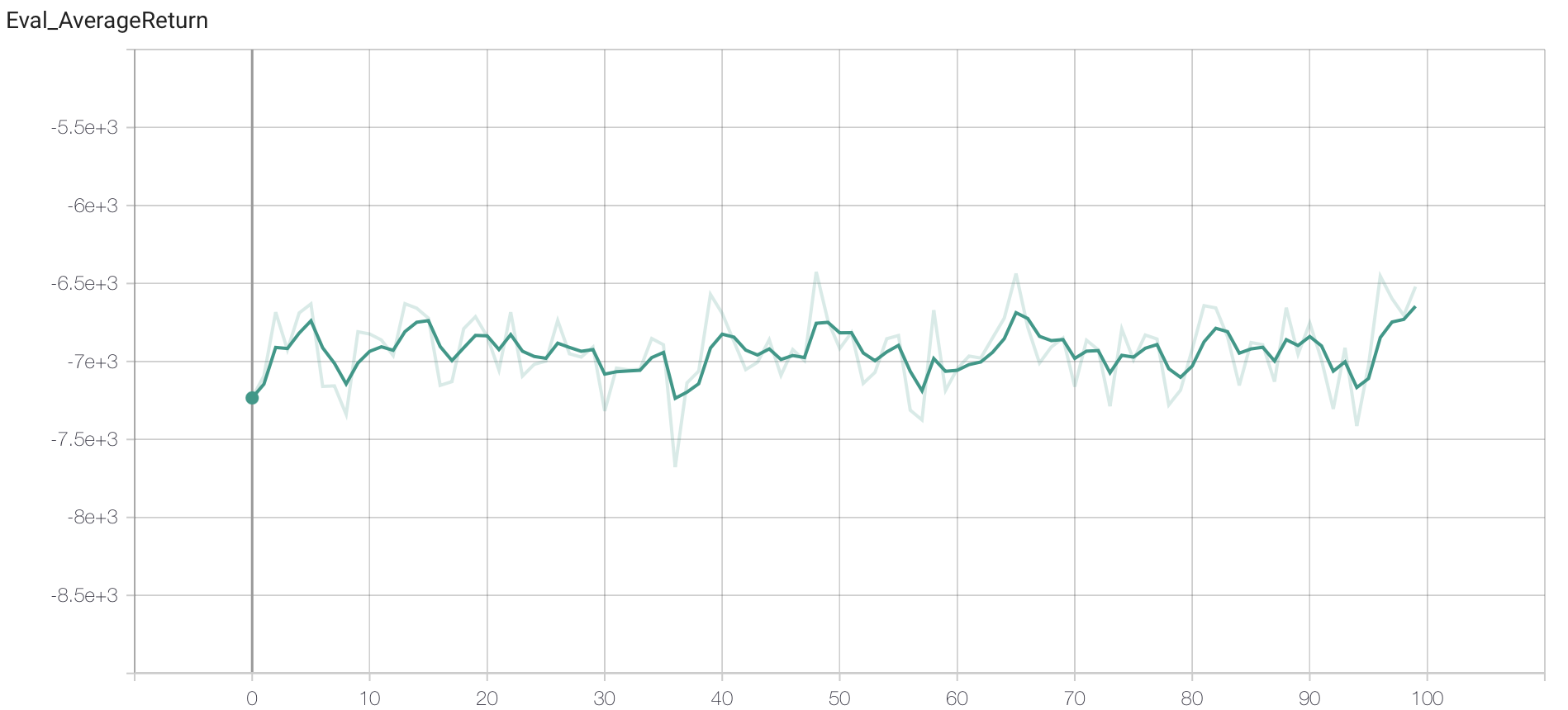}
  \caption{\texttt{eval\_batch\_size = 1000}}
  \label{fig:exp4}
\end{subfigure}
\caption{Performance of CQL agent.}
\label{fig:exp_CQL}
\end{figure*}

\begin{figure*} [ht]
    \centering
    \includegraphics[width=0.75\linewidth]{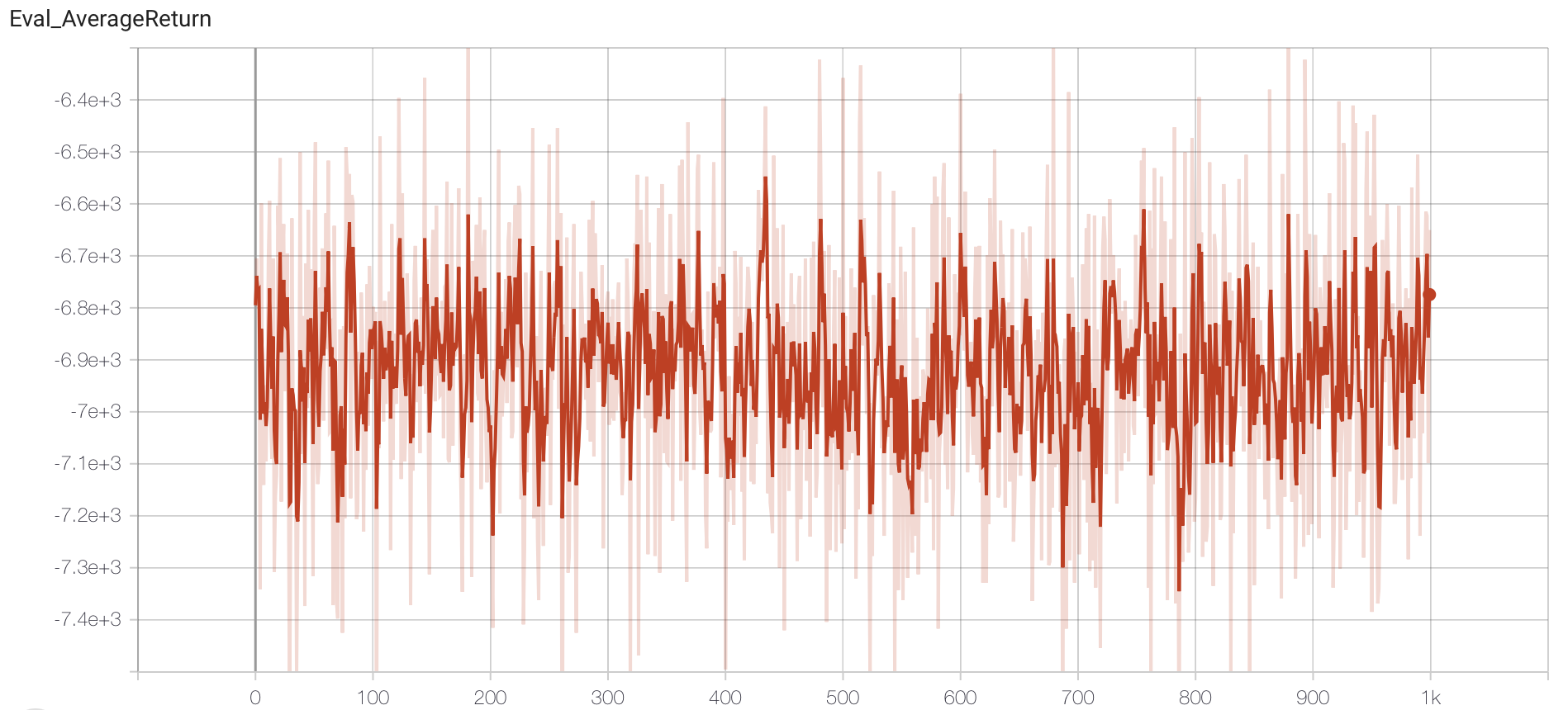}
    \caption{Performance of BC agent (\texttt{n\_ier = 1000}).}
    \label{fig:exp_BC_long}
\end{figure*}

For both BC and CQL, we use the generated offline dataset to train the agent. We have tried various settings of the reinforcement learning parameters, e.g., the neural network structure, the learning rate, the batch size, \texttt{cql\_alpha}, etc. By comparing the learning curves for different \texttt{eval\_batch\_size} (i.e., Figure \ref{fig:exp1} versus \ref{fig:exp2}, Figure \ref{fig:exp3} versus \ref{fig:exp4}), it can be found that the increased \texttt{eval\_batch\_size} will significantly reduce the variance of the model performance and result in more stable learning curves. However, it is also observed that neither of our algorithms can actually start learning to get some improvement in the first $100$ iterations (Figure \ref{fig:exp_BC} \& \ref{fig:exp_CQL}), even in $1000$ iterations (Figure \ref{fig:exp_BC_long}). As such, we analyzed some possible reasons for the failure of our implementation. First, due to time limitation, we only utilized the actual weather data at the destination airport as inputs instead of the forecast weather data. Also, we didn't consider the canceled flights and airport swapping behaviors during GDP implementation as we couldn't find the available database. Second, when developing the $SAGDP\_ENV$, we oversimplified the settings and input information. For the initial design, as each GDP rollout comes from one day during 2019 at EWR with unique weather, operational conditions and flight schedules, we incorporated the large volumes of post processing features of airport and flight data of the whole year as the input when designing the environment. However, the implemented model always uncover bugs and run extremely slow. Instead, we set the environment parameters randomly based on the intrinsic features of the database, which sometimes could be very different from actual situations. Last but not least, the models we selected here might not be appropriate for our application.

\begin{figure*} [ht]
    \centering
    \includegraphics[width=1.0\linewidth]{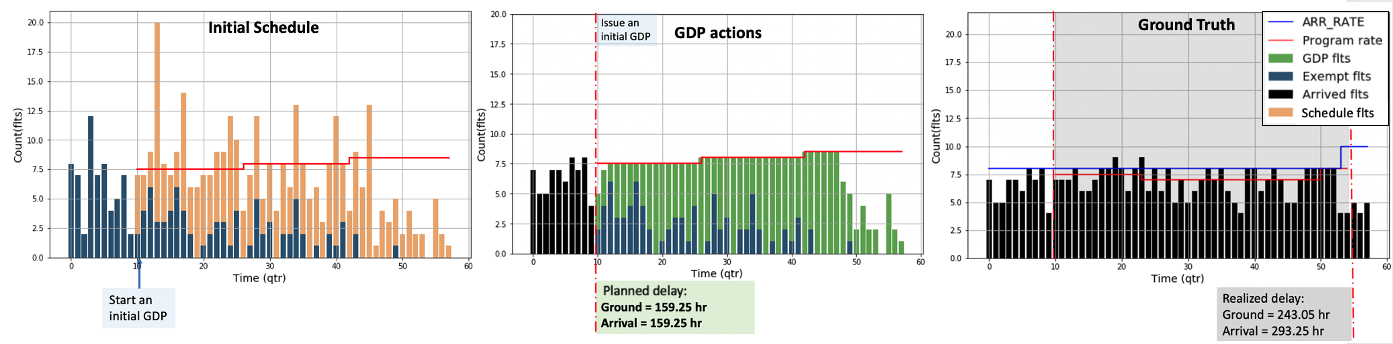}
    \caption{Visualization of BC model results for one GDP }
    \label{fig:paar}
\end{figure*}

To check the model performance in our application, we randomly select one GDP day as an example: 12 hours of operation in Jan $12^{th}$, 2019 at EWR, as shown in Figure \ref{fig:paar}. The GDP was first modeled and released starting from 10:00 am with a duration of 48 quarter hours. From the results of BC model with an evaluation batch size of 100, we get the list of last updated PAAR for these 48 quarter hours. In Figure \ref{fig:paar}, 3 bar plots with different scenarios are presented for the same period of time and each bar represents the count of corresponding flights at given quarter hour. In the first plot, it shows the original scheduled arrival demand for each quarter hour. The red line is the initial PAAR from BC model, we can see the demand excessively exceed the forecast capacity. We detect the GDP in-scope flights if its schedule arrival time is within the GDP time window, and its departure airport is in the restricted control centers and the flight has not taken off when the program is released. Then orange bar is the in-scope flights and blue bars are the exempt flights which are out of control regions or have already taken off. Then the queuing diagram and the RBS algorithm (as described in $SAGDP\_ENV$ in the Scenario Generation section) are applied to assign the slots and ground delay for a GDP event. There will be 159 hours of ground and arrival delay in plan, which indicates all the arrival delay can be shifted to the ground shown in the second graph. In the last graph, we compared with the actual airport capacity and actual throughput at EWR during the same time period. In the actual execution, there are 243 hours of ground delay and 293 hour of arrival delay. The blue line is the actual arrival rate while the red line is the last updated program rate. Based on this example, we can see BC model result underestimate the true airport capacity, and our model at current stage doesn't have ability to predict the performance of GDP after execution. However, the overall performance of BC model and CQL model for different GDP days could vary with the operational and weather conditions, so further evaluation and exploration are needed in the future study.

\section{Conclusions}
In this paper, we applied two reinforcement learning (RL) models, Behavioral Cloning (BC) and Conservative Q-Learning (CQL), to optimize Ground Delay Programs (GDP) at airports, focusing on adjusting airport program rates. These models are part of an initiative to integrate advanced machine learning into air traffic management to minimize delays and enhance efficiency.

We developed a reward function to address key aspects of airport operations—ground and airborne delays, and terminal area congestion, each weighted based on its impact on traffic flow. The simulation environment, $SAGDP_ENV$, was constructed using a queuing diagram and ration-by-schedule algorithm, incorporating real airport capacity, flight demand, weather variability, and operational disruptions.

Using 2019 data from Newark Liberty International Airport (EWR), we aimed to predict the airport's program rate across multiple time intervals. However, the models failed to learn or improve, likely due to the oversimplified assumptions in the RL environment setup that failed to reflect the complexity of actual airport operations.

The limitations of our study are multi-fold and provide several avenues for future research:
\begin{itemize}
    \item[(1)] Comprehensiveness of GDP Modeling: Our current models are limited to optimizing the program rate set two hours ahead of operation. They do not account for other critical GDP parameters such as scope and duration, nor do they consider airline operational strategies like flight cancellations and substitutions during GDP implementations. Future iterations of the model should aim to incorporate these aspects into the RL framework to provide a more holistic approach to traffic flow management.
    \item[(2)] Inclusion of Weather Data: One significant oversight was the exclusion of forecast and enroute weather data in our models. These elements are crucial as they significantly influence GDP decisions and effectiveness. Integrating comprehensive weather forecasting into the simulation environment could greatly enhance the predictive accuracy and operational relevance of the models.
\end{itemize}


\bibliographystyle{IEEEtran}
\bibliography{refs.bib}
\vspace{12pt}
\end{document}